\documentclass{article}
\usepackage{spconf,amsmath,amssymb,amsfonts,graphicx}
\usepackage{cite}
\usepackage{xcolor}

\usepackage{pgfplots}
\pgfplotsset{compat=1.18}
\usepackage{booktabs}
\usepackage{multirow}
\usepackage{subcaption}
\usepackage{arydshln}
\usepackage{hyperref}
\usepackage{cleveref}
\usepackage{pifont}
\newcommand{\cmark}{{\color{green}\ding{51}}}%
\newcommand{\xmark}{{\color{red}\ding{55}}}%

\usepackage{colortbl}
\definecolor{ownrow}{gray}{0.95}
\definecolor{plotblue}{RGB}{31,119,180}

\let\oldthebibliography\thebibliography
\renewcommand{\thebibliography}[1]{%
  \oldthebibliography{#1}%
  \footnotesize
  \setlength{\itemsep}{4pt plus 0.5pt}%
  \setlength{\parskip}{0pt}%
}

\title{Enriching Speech Emotion Representations with Conversational Context}

\name{Arthur Peuvot$^{\star}$ \qquad Romaric Besan\c{c}on$^{\star}$ \qquad Ga\"el de Chalendar$^{\star}$ \qquad Bianca Vieru$^{\star}$ \qquad Ioana Vasilescu$^{\dagger}$
\thanks{This publication was made possible by the use of the CEA List FactoryIA supercomputer, financially supported by the Ile-de-France Regional Council. This work was partially supported by the ANR-23-PEIA0008 SHARP project in the context of the France 2030 program.}}
\address{$^{\star}$~Universit\'e Paris-Saclay, CEA, List, France\\
\texttt{\{arthur.peuvot, romaric.besancon, gael.de-chalendar, bianca.vieru\}@cea.fr}\\
$^{\dagger}$~LISN, CNRS, Universit\'e Paris-Saclay, France\\
\texttt{ioana.vasilescu@lisn.fr}}

\begin{document}
\ninept
\maketitle
%
\begin{abstract}
Detecting emotions is necessary for building systems that can accurately and adaptively interact with humans. Speech Emotion Recognition (SER) has become an important research focus to develop intelligent spoken interfaces. However, most studies predict emotions at the utterance level, ignoring the conversational context, along with the emotional flow and speaker interactions it carries. In this paper, we introduce ACERT (Averaged Contextual Emotion Representation through Time), a module that integrates a flexible-length window of conversational context to better capture emotional evolution in spoken interactions. To evaluate the robustness of this method, we conducted experiments on datasets spanning diverse emotionally expressive styles and contexts. ACERT outperforms current state-of-the-art (SOTA) approaches on IEMOCAP, establishes the first context-aware benchmark on SAFE, and obtains strong results on MELD for unweighted, class-balanced metrics. Ablation studies show that ACERT's gains come from emotional and conversational continuity, rather than from speaker identity or acoustic conditions.
\end{abstract}

\section{Introduction}

Natural human interaction requires adaptive strategies that dynamically link past and present conversational elements, anticipate interlocutors’ reactions, and integrate multi-level inputs, from acoustic cues to complex emotional states. Humans excel at this adaptability, underpinned by several foundational cognitive frameworks such as Theory of Mind, concerned with the anticipation of others’ mental states and emotional responses \cite{lake2017building}, pragmatic inference to detect implicatures \cite{grice1975logic}, or socio-cultural norms as filters for emotion expression \cite{evidence_laukka_2014}. In contrast, current computational models for Speech Emotion Recognition (SER) predominantly rely on statistical patterns and probabilistic predictions, often failing to capture the depth and contextual fluidity of human understanding. Our goal is to build models that use conversational context to predict upcoming emotional states for more adaptive responses, and maintain high performance across diverse interactional contexts.

With the emergence of the Transformer architecture \cite{vaswani2017attention}, the field of SER has recently transitioned from hand-crafted acoustic features to deep learning representations. Self-Supervised Learning (SSL) models such as wav2vec 2.0 \cite{baevski2020wav2vec}, HuBERT \cite{hsu2021hubert}, and WavLM \cite{chen2022wavlm} have set new baselines by extracting rich features from raw audio that capture both content and prosody \cite{chen2023exploring}. To further specialize these architectures, recent works have focused on emotional adaptation. For instance, emotion2vec \cite{ma2024emotion2vec} utilizes self-supervised online distillation on unlabeled emotional data, while ExHuBERT \cite{amiriparian2024exhubert} enhances the HuBERT backbone through layer duplication and fine-tuning on emotionally rich datasets, ensuring adaptability for affective tasks.

Using these SSL models, recent works extract structural nuances of speech through diverse strategies. Some approaches prioritize hierarchical modeling, such as SpeechFormer++ \cite{chen2023speechformer++} and MSTR \cite{li2024multi}, which mimic the relationship between frames, phones, and words to balance fine and coarse-grained information. Others focus on temporal dynamics at the utterance level. DWFormer \cite{chen2023dwformer} uses dynamic window splitting to capture local importance, while \cite{wang2025speech} employ Neural Controlled Differential Equations to model high-dimensional time-series features. To ensure that the captured signal is purely affective, DTNet \cite{yuan2024disentanglement} introduces an identity-aware module to disentangle emotional features from speaker-specific acoustic traits.

The majority of SER datasets contain isolated or read utterances \cite{livingstone2018ryerson, cao2014crema, burkhardt2005database, banziger2007using, moine2020att}. Thus, most SER approaches still predict emotions at the utterance level, not taking into account the intra and inter-speaker emotional dynamics inherent in dialogue. To address this, some multimodal works like \cite{hazarika2018icon, poria2017context, shi2023emotion} attempt to model intra and inter-speaker emotional influences. Among audio-based approaches, CHAN \cite{tellai2024novel} uses the immediately preceding utterance from the speaker and their interlocutor in a dyadic setting, and ESA CRF \cite{chen2022emotion} models transitions between consecutive utterance-level emotions. However, these approaches remain limited. CHAN restricts context to a single preceding utterance, while ESA CRF operates on the sequence of utterance-level predictions rather than on the underlying audio representations. Multimodal approaches require textual or multi-speaker information that is not always available.

In contrast, we introduce a simple and effective method that enriches the target utterance's audio representation with a flexible-length window of preceding context through mean pooling. This approach relies on the audio modality alone and operates regardless of the speaker. It outperforms the state of the art on IEMOCAP \cite{busso2008iemocap}, while establishing strong results on SAFE \cite{clavel2006safe} and MELD \cite{poria2019meld}.

The main contributions of this work are as follows:
\begin{enumerate}
    \item \textbf{ACERT (Averaged Contextual Emotion Representation through Time):} A module designed to enrich utterance representations by integrating conversational information. It aggregates the preceding context through temporal mean pooling and fuses it with the target utterance representation, bridging the gap between isolated audio analysis and dynamic dialogue. The source code for ACERT is publicly available\footnote{\url{https://github.com/apeuvot/ACERT}}.
    \item \textbf{Evaluation on diverse interactional dynamics:} We evaluate our approach on three datasets with distinct interactional dynamics and emotionally expressive styles: IEMOCAP (dyadic scripted or improvised dialogues), MELD (multi-party conversations from the \textit{Friends} TV show), and SAFE (intense emotional interactions from movies such as thrillers).
\end{enumerate}

\section{Methodology}

ACERT is a block that can be added on top of any feature extractor, from SSL models to other SOTA SER methods. It aggregates the conversational context through temporal mean pooling and fuses it with the target utterance's features. The resulting context-augmented representation is then fed into a classification head.

\subsection{Problem setting}

A conversation is defined as an ordered sequence of utterances, $\mathcal{U} = \{u_1, u_2, \dots, u_N\}$, where $N$ denotes the total number of utterances in the conversation and $u_i$ represents the $i$-th utterance.
Each utterance $u_i$ is associated with a speech signal $x_i \in \mathbb{R}^{T_i}$ and an emotion label $y_i \in \{1, \dots, C\}$, where $C$ is the number of emotion classes. The objective of the SER task is to learn a mapping function $f: \mathbb{R}^{T_i} \to \{1, \dots, C\}$ that assigns an emotion $y_i$ to each speech signal $x_i$.

\subsection{ACERT}

\noindent\textbf{Context definition.} Let $t \in \mathbb{R}^+$ denote a fixed temporal context window, defined as a hyperparameter of the system and corresponding to the duration of context alone, excluding the target utterance. For each target utterance $u_i$, an input audio segment $x^{(i)}$ is constructed by concatenating $u_i$ with as many preceding utterances as fit within $t$ seconds of context, so that the total duration of $x^{(i)}$ equals the duration of $u_i$ plus $t$. If the available preceding utterances exceed $t$ seconds, the earliest of these context utterances is truncated to fit the context window.

\noindent\textbf{Context design choices.} To define this context design, we ran preliminary experiments comparing several configurations. Using both preceding and following context performed comparably to preceding-only context (\Cref{tab:comparison_sota}). We therefore use only preceding-utterance context, as it also avoids the latency of waiting for future utterances in a real use case. Similarly, integrating speaker-dependent context, obtained either from ground-truth speaker labels or from a speaker diarization step using k-means, demonstrated no statistically significant difference in performance compared to speaker-independent context (\Cref{tab:comparison_sota}). We therefore include context regardless of the speaker, which also avoids the risk of diarization errors propagating to the SER stage. As speaker information is not available in all datasets, this speaker-independent, preceding-only context design further ensures consistent evaluation across corpora.

\noindent\textbf{Feature extraction.} The input audio segment $x^{(i)}$, containing the target utterance together with its $t$-second preceding context, is processed by a feature extractor and then passed through a learnable linear layer with ReLU activation to obtain frame-level representations $\mathbf{H}$.

\noindent\textbf{Enrichment mechanism.} The ACERT block incorporates information from the preceding utterances into the target utterance's features. The frame-level features $\mathbf{H}$ are split into target frames $\mathbf{H}_t$ and context frames $\mathbf{H}_c$. The context frames are aggregated through temporal mean pooling:
\begin{equation}
    \mathbf{c} = \frac{1}{|\mathbf{H}_c|}\sum_{h \in \mathbf{H}_c} h
\end{equation}
The pooled vector $\mathbf{c}$ is added to every target frame, followed by a feed-forward network with residual connections and LayerNorm \cite{vaswani2017attention}:
\begin{equation}
    \mathbf{H}_t' = \text{LN}(\mathbf{H}_t + \mathbf{c}), \quad \mathbf{H}_t'' = \text{LN}\big(\mathbf{H}_t' + \text{FFN}(\mathbf{H}_t')\big)
\end{equation}
where $\text{FFN}(\cdot)$ is a two-layer feed-forward network with ReLU activation. Temporal average pooling over $\mathbf{H}_t''$ produces an utterance-level representation $\bar{\mathbf{h}}_i$, passed through a classification head (a ReLU activation followed by a linear layer) to predict one of the $C$ emotion classes.

\begin{table}[t]
    \centering
    \small
    \setlength{\tabcolsep}{4pt}
    \caption{Statistical summary of the IEMOCAP, SAFE, and MELD datasets.}
    \label{tab:dataset_stats}
    \begin{tabular}{lccc}
        \toprule
        \textbf{Statistic} & \textbf{IEMOCAP} & \textbf{SAFE} & \textbf{MELD} \\
        \midrule
        \multicolumn{4}{l}{\textit{Utterance level}} \\
        Total files          & 5531   & 5638  & 13706 \\
        Mean duration (s)    & 4.55   & 4.28  & 3.19  \\
        Median duration (s)  & 3.58   & 2.72  & 2.49  \\
        \midrule
        \multicolumn{4}{l}{\textit{Conversation level}} \\
        Total conversations  & 151    & 398   & 1432  \\
        Mean duration (s)    & 166.63 & 60.56 & 30.52 \\
        Median duration (s)  & 171.36 & 52.88 & 27.52 \\
        \midrule
        \multicolumn{4}{l}{\textit{Emotional stability}} \\
        Mean consec. same emotion utt. & 7.38 & 5 & 1.71 \\
        Mean same emotion duration (s) & 33.6 & 27.5 & 5.5 \\
        \bottomrule
    \end{tabular}
\end{table}

\section{Experimental setup}

\subsection{Datasets}

Finding real-life corpora that are freely available for use, particularly those capturing emotions in naturalistic conditions, remains a significant challenge. Most publicly available datasets are limited in terms of emotional depth or context. Our method, based on conversational context, requires datasets that include entire conversations to capture the dynamic emotional flow. To address this, we propose a preliminary evaluation using acted corpora, which, despite being scripted, offer diverse acquisition contexts and emotionally expressive styles. The selected corpora are IEMOCAP, MELD and SAFE. These corpora have almost no missing utterances guaranteeing a reliable continuity in contextual flow. Although some of these corpora provide additional modalities (e.g., text or video), ACERT only uses the audio modality.

The IEMOCAP corpus is widely used in SER. It consists of 12 hours of dyadic conversations performed by 10 professional actors during improvised or scripted interactions. We used the same emotion labels as most SOTA methods: happiness (merged with excitement), anger, sadness, and neutrality.

MELD contains 13.7 hours of conversations from the TV show \textit{Friends}. In contrast to IEMOCAP, MELD introduces multi-party dynamics characterized by an over-acted and emotionally dense style, typical of sitcoms, with numerous emotional shifts. The emotional categories are: anger, disgust, fear, joy, neutrality, sadness, and surprise.

The Situation Analysis in a Fictional and Emotional (SAFE) corpus is designed to analyze strong emotions in extreme contexts such as natural disasters and physical or psychological threats and aggression. It was created by extracting 400 scenes from 30 different movies. It provides about 7 hours of audio with 400 different speakers. The emotional labels are: fear, other negative emotions, positive emotions, and neutrality.

Detailed statistics regarding utterance and conversation lengths for all datasets are provided in the \Cref{tab:dataset_stats}.

\suppressfloats[t]

\subsection{Training and evaluation procedure}
\label{subsec:training_procedure}

We use HuBERT-large\footnote{\url{https://huggingface.co/facebook/hubert-large-ll60k}} as the feature encoder. Models are trained using a cross-entropy loss function, a batch size of 16 and a learning rate of $1.3 \times 10^{-4}$. The evaluations are conducted using a context window $t \in \{5k \mid k \in \{1, \dots, 8\}\}$.

For IEMOCAP and SAFE, we evaluate our method in a speaker-independent 5-fold cross-validation setting. For IEMOCAP, we use the standard evaluation method used in SOTA works (e.g., \cite{chen2023dwformer, chen2023exploring}), where each test fold contains one of the 5 sessions. For SAFE each test fold consists of scenes from 6 of the 30 movies in the dataset. For MELD, we use the standard training, development, and test splits \cite{chen2023speechformer++, chen2023dwformer, li2024multi}. Each experiment is run 5 times.

We report performance using Unweighted Accuracy (UA), Weighted Accuracy (WA), Macro-F1, and Weighted-F1 scores along with their respective standard deviations.

\subsection{Ablation studies}
\label{subsec:ablation_setup}

To test whether ACERT's gains depend on conversational and emotional continuity, and not just on speaker identity or recording conditions, we design three ablation studies on IEMOCAP at the optimal context window $t=25\,$s. In the \textit{same session, continuous context} condition, the true preceding context is replaced by a continuous segment from a single other conversation in the same session (same speakers, same recording conditions), preserving turn-taking but unrelated to the target conversation. In the \textit{same session, random context} condition, it is instead replaced by a random mixture of utterances from all other conversations in the same session, with no temporal or conversational continuity. Comparing these two conditions isolates the effect of the context's continuity, independently of its relevance to the target conversation. Since both conditions control for speaker identity and recording conditions, any drop in performance can be attributed to the loss of conversational and emotional continuity. Finally, in the \textit{same conversation, shuffled order} condition, the preceding context is kept but its utterances are presented in random order, isolating the effect of temporal order alone, independently of context relevance.

\captionsetup[subfigure]{aboveskip=2pt, belowskip=0pt}
\captionsetup{aboveskip=3pt}

\captionsetup[subfigure]{aboveskip=2pt, belowskip=0pt}
\captionsetup{aboveskip=3pt}

\begin{figure}[t]
    \centering
    \begin{minipage}[c]{0.06\linewidth}
        \centering
        \rotatebox{90}{\small Unweighted Accuracy (\%)}
    \end{minipage}%
    \begin{minipage}[c]{0.92\linewidth}
        \centering
        \begin{subfigure}[b]{\linewidth}
            \centering
            \begin{tikzpicture}
                \begin{axis}[
                    width=0.95\linewidth,
                    height=2.6cm,
                    title={Context window length (s)},
                    axis x line*=bottom,
                    axis y line*=left,
                    grid=major,
                    grid style={dashed, gray!25},
                    ymajorgrids=true,
                    xmajorgrids=false,
                    xmin=-1, xmax=40,
                    ymin=67, ymax=82,
                    ytick={70,75,80},
                    tick label style={font=\scriptsize},
                    title style={font=\small},
                ]
                \addplot[
                    color=plotblue, thick,
                    mark=*, mark size=1.8pt,
                    mark options={fill=plotblue, draw=white, line width=0.4pt},
                    error bars/.cd, y dir=both, y explicit
                ] table[x=time, y=UAR, y error=UAR_std, col sep=comma] {results/uar_iemocap.csv};
                \end{axis}
            \end{tikzpicture}
            \caption{IEMOCAP}
            \label{fig:uar_iemocap}
        \end{subfigure}

        \begin{subfigure}[b]{\linewidth}
            \centering
            \begin{tikzpicture}
                \begin{axis}[
                    width=0.95\linewidth,
                    height=2.6cm,
                    axis x line*=bottom,
                    axis y line*=left,
                    grid=major,
                    grid style={dashed, gray!25},
                    ymajorgrids=true,
                    xmajorgrids=false,
                    xmin=-1, xmax=40,
                    ymin=40, ymax=46,
                    ytick={40,43,46},
                    tick label style={font=\scriptsize},
                ]
                \addplot[
                    color=plotblue, thick,
                    mark=*, mark size=1.8pt,
                    mark options={fill=plotblue, draw=white, line width=0.4pt},
                    error bars/.cd, y dir=both, y explicit
                ] table[x=time, y=UAR, y error=UAR_std, col sep=comma] {results/uar_safe.csv};
                \end{axis}
            \end{tikzpicture}
            \caption{SAFE}
            \label{fig:uar_safe}
        \end{subfigure}

        \begin{subfigure}[b]{\linewidth}
            \centering
            \begin{tikzpicture}
                \begin{axis}[
                    width=0.95\linewidth,
                    height=2.6cm,
                    axis x line*=bottom,
                    axis y line*=left,
                    grid=major,
                    grid style={dashed, gray!25},
                    ymajorgrids=true,
                    xmajorgrids=false,
                    xmin=-1, xmax=40,
                    ymin=10, ymax=30,
                    ytick={10,20,30},
                    tick label style={font=\scriptsize},
                ]
                \addplot[
                    color=plotblue, thick,
                    mark=*, mark size=1.8pt,
                    mark options={fill=plotblue, draw=white, line width=0.4pt},
                    error bars/.cd, y dir=both, y explicit
                ] table[x=time, y=UAR, y error=UAR_std, col sep=comma] {results/uar_meld.csv};
                \end{axis}
            \end{tikzpicture}
            \caption{MELD}
            \label{fig:uar_meld}
        \end{subfigure}
    \end{minipage}
    \caption{Unweighted Accuracy (\%) as a function of the context window length (s) on IEMOCAP, SAFE, and MELD.}
    \label{fig:uar_all_datasets}
\end{figure}
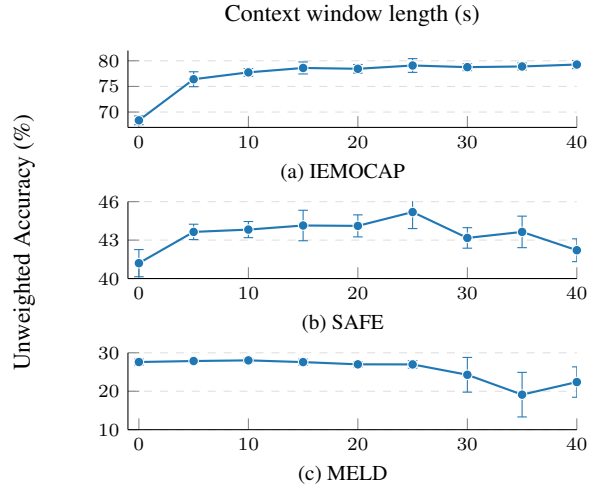

\section{Results and discussion}

\subsection{Experimental results}
\label{subsec:exp_results}

\Cref{fig:uar_iemocap} shows the results of our method on IEMOCAP. The performance improves significantly as the context window length increases before eventually reaching a plateau. This displays the effectiveness of using context rather than relying solely on the target utterance, which lasts on average 4.55\,s (see \Cref{tab:dataset_stats}), for prediction. Adding only $t=5\,$s of context already improves UA by 8.02 points over the context-free baseline (from 68.38\,\% to 76.40\,\%), and the highest overall performance is achieved at $t=25\,$s with an UA of 79.08\,\%, a gain of 10.70 points.

As with IEMOCAP, we observe a significant increase in performance with context on the SAFE dataset (\Cref{fig:uar_safe}), peaking at $t=25\,$s with a 3.99\,\% point gain in UA over the context-free baseline (from 41.20\,\% to 45.19\,\% UA), before declining as the context window is extended further.

In contrast, for MELD (\Cref{fig:uar_meld}) we observe no statistically significant improvement of the results when using a short context window. The reasons behind this stagnation are discussed in \Cref{subsec:limitations}.

\begin{table*}[t]
    \centering
    \caption{Comparison to baseline and current SOTA methods on IEMOCAP, SAFE, and MELD. Preliminary design experiments and ablation studies on IEMOCAP. Rows shaded in gray indicate our method.}
    \begin{tabular}{llccccc}
        \toprule
        \textbf{Dataset} & \multicolumn{2}{l}{\textbf{Method} \hspace{1.5cm} \textbf{Use conversational context}} & \textbf{UA (\%)} & \textbf{WA (\%)} & \textbf{Macro-F1 (\%)} & \textbf{Weighted-F1 (\%)}\\
        \midrule
        \multirow{20}{*}{IEMOCAP}
        & HuBERT-large fine-tuning (baseline) & \xmark & 68.38 ± 0.85 & 67.03 ± 0.80 & 67.30 ± 0.87 & 66.41 ± 0.88 \\
        & DWFormer (2023) \cite{chen2023dwformer} & \xmark & 73.9 & 72.3 & - & - \\
        & P-TAPT (2023) \cite{chen2023exploring} & \xmark & 74.3 & - & - & - \\
        & DTNet (2024) \cite{yuan2024disentanglement} & \xmark & 74.8 & - & - & - \\
        & NCDEs-Classifier (2025) \cite{wang2025speech} & \xmark & 74.19 & 73.37 & - & - \\
        & Zaho et al. (2025) \cite{zhao2025knowledge} & \xmark & 71.27 & 70.25 & - & - \\ 
        \noalign{\vskip 2pt}
        & ESA CRF (2022) \cite{chen2022emotion} & \cmark &  74.47 & 73.17 & - & - \\
        & Shi et al. (2023), speech only \cite{shi2023emotion} & \cmark & 65.01 & - & 65.91 & - \\
        & CHAN (2024) \cite{tellai2024novel} & \cmark & 75.9 & 75 & - & - \\
        & \cellcolor{ownrow}ACERT (ours), t = 25\,s & \cellcolor{ownrow}\cmark & \cellcolor{ownrow}\textbf{79.08} ± 1.34 & \cellcolor{ownrow}\textbf{77.83} ± 0.88 & \cellcolor{ownrow}\textbf{78.22} ± 1.01 & \cellcolor{ownrow}\textbf{77.64} ± 0.91\\[2pt]
        \cline{2-6}
        \noalign{\vskip 2pt}
        & \multicolumn{6}{l}{\textbf{\textit{ACERT's preliminary design experiments (t = 25\,s)}}} \\
        & Preceding and following context & \cmark & 79.00 ± 0.81 & 77.91 ± 0.69 & 78.32 ± 0.74 & 77.69 ± 0.7 \\
        & Speaker-dependent (ground truth) & \cmark & 79.73 ± 1.02 & 78.71 ± 0.89 & 79.07 ± 0.81 & 78.46 ± 0.80 \\
        & Speaker-dependent (k-means diarization) & \cmark & 78.53 ± 0.58 & 77.2 ± 0.63 & 77.64 ± 0.55 & 76.87 ± 0.72 \\ 
        \noalign{\vskip 2pt}
        & \multicolumn{6}{l}{\textbf{\textit{ACERT's ablation studies (t = 25\,s)}}} \\
        & Same session, continuous context & \cmark & 69.82 ± 1.11 & 68.54 ± 1.28 & 68.63 ± 1.34 & 68.12 ± 1.33 \\
        & Same session, random context & \cmark & 70.09 ± 0.51 & 68.98 ± 0.49 & 69.22 ± 0.5 & 68.69 ± 0.46 \\
        & Same conversation, shuffled order & \cmark & 75.87 ± 1.10 & 75.02 ± 0.80 & 75.15 ± 0.87 & 74.77 ± 0.79 \\
        \midrule
        \multirow{4}{*}{SAFE}
        & HuBERT-large fine-tuning (baseline) & \xmark & 41.20 ± 1.06 & 48.54 ± 0.48 & 40.39 ± 0.94 & 48.09 ± 0.37 \\
        \noalign{\vskip 2pt}
        & \cellcolor{ownrow}ACERT (ours), t = 25\,s & \cellcolor{ownrow}\cmark & \cellcolor{ownrow}\textbf{45.19} ± 1.29 & \cellcolor{ownrow}\textbf{51.42} ± 1.24 & \cellcolor{ownrow}\textbf{43.59} ± 1.36 & \cellcolor{ownrow}\textbf{50.88} ± 1.47 \\
        \midrule
        \multirow{10}{*}{MELD}
        & HuBERT-large fine-tuning (baseline) & \xmark & 27.60 ± 0.80 & 49.83 ± 1.17 & 28.49 ± 0.92 & 47.05 ± 0.62 \\
        & SpeechFormer++ (2023) \cite{chen2023speechformer++} & \xmark & 27.3 & 51 & - & 47 \\
        & DWFormer (2023) \cite{chen2023dwformer} & \xmark & - & - & - & 48.5 \\
        & MSTR (2024) \cite{li2024multi} & \xmark & - & - & - & 46.15 \\
        & emotion2vec (2024) \cite{ma2024emotion2vec} & \xmark & - & 51.88 & 28.03 & \textbf{48.7} \\
        & Zaho et al. (2025) \cite{zhao2025knowledge} & \xmark & - & \textbf{52.28} & - & 46.96 \\
        \noalign{\vskip 2pt}
        & Shi et al. (2023), speech only \cite{shi2023emotion} & \cmark & \textbf{28.67} & - & 25.97 & - \\
        & \cellcolor{ownrow}ACERT (ours), t = 10\,s & \cellcolor{ownrow}\cmark & \cellcolor{ownrow}28.03 ± 0.58 & \cellcolor{ownrow}47.19 ± 2.06 & \cellcolor{ownrow}\textbf{28.62} ± 0.80 & \cellcolor{ownrow}45.64 ± 1.41 \\
        \bottomrule
    \end{tabular}
    \label{tab:comparison_sota}
\end{table*}

\subsection{Comparison with baseline and SOTA works}

\Cref{tab:comparison_sota} compares ACERT against the baseline and SOTA methods, with and without conversational context. For the SOTA approaches, values are taken from the original papers. For ACERT, we report the results of the experiments with the context window $t$ that achieved the highest UA.

On IEMOCAP, ACERT outperforms not only the context-free baseline and SOTA methods without context, but also all context-aware methods, including CHAN, the strongest context-aware baseline, by 3.18 points in UA and 2.83 points in WA. 

On SAFE, no prior context-aware evaluation exists but ACERT improves over the context-free baseline by 3 to 4 points across all four metrics, establishing the first context-aware benchmark on this dataset. 

On MELD, ACERT matches or exceeds \cite{shi2023emotion}, the other context-aware method, in UA and Macro-F1, achieving the best Macro-F1 across all methods regardless of context. It also outperforms the context-free baseline on both metrics, though it still falls below on weighted metrics.

\subsection{Ablation studies}
\label{subsec:ablation_results}

\Cref{tab:comparison_sota} reports the results of the three ablation studies described in \Cref{subsec:ablation_setup}. All three perform notably worse than ACERT (69.82\,\%, 70.09\,\% and 75.87\,\% UA against 79.08\,\%), despite using context from the same speakers and recording conditions. This gap holds consistently across all four metrics. These results confirm that the improvement brought by ACERT does not simply come from the presence of additional speech of the current speakers or from matching acoustic conditions, but specifically from the conversational and emotional continuity of the context. Moreover, the two unrelated-context studies perform similarly to each other (69.82\,\% vs. 70.09\,\% UA, within one standard deviation), suggesting that once the context is unrelated to the target conversation, its continuity has little additional impact.

Interestingly, both ablation results also remain close to the context-free baseline (68.38\,\% UA), which can itself be regarded as a third, more extreme ablation of ACERT providing no context at all. This indicates that an unrelated context brings almost no benefit over having no context at all, an effect we also observe for MELD in \Cref{subsec:limitations}. In contrast, shuffling the order of the true context gives an intermediate result (75.87\,\% UA), confirming that the temporal order of a relevant context also contributes to ACERT's gains, beyond its relevance alone.

\subsection{Limitations}
\label{subsec:limitations}

As shown in \Cref{subsec:exp_results}, using conversational context fails to improve performance on the MELD dataset. Several factors may explain this finding. First, the number of speakers in each conversation: while IEMOCAP involves two easily distinguishable speakers (one male and one female), MELD averages 3.4 speakers per conversation (SAFE lacks sufficient speaker-related data for a reliable comparison). Furthermore, MELD exhibits a much lower emotional stability than IEMOCAP and SAFE (\Cref{tab:dataset_stats}), with an average same emotion duration of only 5.5\,s, compared to 33.6\,s for IEMOCAP and 27.5\,s for SAFE.
This emotional instability, together with our IEMOCAP ablation study (\Cref{subsec:ablation_results}) showing that context whose emotional state differs from the target utterance provides no benefit over having no context at all, explains why extending the context window does not improve performance on MELD.

\section{Conclusion}

This paper introduces a novel module that leverages conversational context through temporal mean pooling to enrich the representation of a target utterance with information from the surrounding dialogue. Evaluated across three datasets spanning distinct interactional dynamics, ACERT largely surpasses state-of-the-art methods on IEMOCAP and achieves solid results on SAFE and MELD. Our results further show that using context is most effective when conversations are emotionally stable such as in IEMOCAP and SAFE.

Future work will explore ways to automatically detect when context is emotionally relevant to the target utterance, rather than treating all context equally, which could help on datasets with rapid emotional shifts such as MELD. It could also evaluate cross-corpus generalization, for example by training on one dataset and testing on another, to assess whether the benefit of conversational context transfers across different interactional styles.

\bibliographystyle{IEEEbib}
\bibliography{mybib}

\end{document}